\documentclass[10pt,twocolumn,letterpaper]{article}

\usepackage{cvpr}      % To produce the REVIEW version
\definecolor{cvprblue}{rgb}{0.21,0.49,0.74}
\usepackage[pagebackref,breaklinks,colorlinks,allcolors=cvprblue]{hyperref}
\usepackage{amsmath, amssymb, amsthm, bm}
\usepackage[table]{xcolor}
\usepackage{graphicx}
\usepackage{tikz}
\usepackage{amsthm} % 推荐加载 amsthm 宏包以获得更好的定理类环境支持

\usepackage[mathscr]{eucal}
\usepackage{makecell}
\usepackage{multirow}
\usepackage{amsmath}    % 提供 \mathbf, \text, 多行公式等核心数学支持
\usepackage{amssymb}    % 提供额外数学符号（如 \mathbb, \mathcal 等）
\usepackage{amsfonts}   % 通常与 amssymb 配合使用
\usepackage{mathtools}  % amsmath 的增强版，可选但推荐
\usepackage{xparse}     % 用于定义带可选参数或复杂参数的命令（如 \Inputs）
\usepackage{algorithm}
\usepackage{algpseudocode} % Use algorithmicx package
\usepackage{arydshln}
\definecolor{highlightcolor}{RGB}{255, 228, 196}
\algnewcommand{\Inputs}[1]{%
  \State \textbf{Inputs:}
  \Statex \hspace*{\algorithmicindent}\parbox[t]{.8\linewidth}{\raggedright #1}
}
\algnewcommand{\Outputs}[1]{%
  \State \textbf{Outputs:}
  \Statex \hspace*{\algorithmicindent}\parbox[t]{.8\linewidth}{\raggedright #1}
}
\algnewcommand{\Hyperparams}[1]{%
  \State \textbf{Hyperparameters:}
  \Statex \hspace*{\algorithmicindent}\parbox[t]{.8\linewidth}{\raggedright #1}
}
\definecolor{cvprblue}{rgb}{0.21,0.49,0.74}
\usepackage[pagebackref,breaklinks,colorlinks,allcolors=cvprblue]{hyperref}

\def\paperID{3504} % *** Enter the Paper ID here
\def\confName{CVPR}
\def\confYear{2026}

\title{ One Token, Two Fates: A Unified Framework via Vision Token Manipulation Against MLLMs Hallucination }

\author{
Zhan Fa$^{1}$,\quad Yue Duan$^{1}$,\quad  Jian Zhang$^{1}$,\quad  Lei Qi$^{2}$,\quad  Yinghuan Shi$^{1}$\thanks{ Corresponding author.}\\
$^{1}$National Key Laboratory for Novel Software Technology, Nanjing University, China\\
$^{2}$School of Computer Science and Engineering, Southeast University, China\\
{\tt\small \{fazhan, yueduan\}@smail.nju.edu.cn, qilei@seu.edu.cn, \{zhang.jian,syh\}@nju.edu.cn}
}

\begin{document}
\maketitle

\begin{abstract}
   Current training-free methods tackle MLLM hallucination with separate strategies: either enhancing visual signals or suppressing text inertia. However, these separate methods are insufficient due to critical trade-offs: simply enhancing vision often fails against strong language prior, while suppressing language can introduce extra image-irrelevant noise. Moreover, we find their naive combination is also ineffective, necessitating a \textbf{unified framework}. We propose such a framework by focusing on the core asset: the vision token. Our design leverages two key insights: (1) augmented images offer complementary visual semantics, and (2) removing vision tokens (information-gap) isolates hallucination tendencies more precisely than distorting images (modality-gap). Based on these, our framework uses vision tokens in two distinct ways, both operating on latent representations: our Synergistic Visual Calibration (SVC) module incorporates augmented tokens to strengthen visual representations, while our Causal Representation Calibration (CRC) module uses pruned tokens to create latent-space negative samples for correcting internal model biases. By harmonizing these two roles, our framework effectively restores the vision-language balance, significantly reducing object hallucinations, improving POPE accuracy by an average of 2\% absolute on LLaVA-1.5 across multiple benchmarks with only a 1.06x inference latency overhead. 
\end{abstract}

\section{Introduction}
\label{sec:intro}

%----------- NEW INTRO FIGURE (intro.jpg) -----------
\begin{figure}[t]
 \centering
 \includegraphics[width=0.95\linewidth]{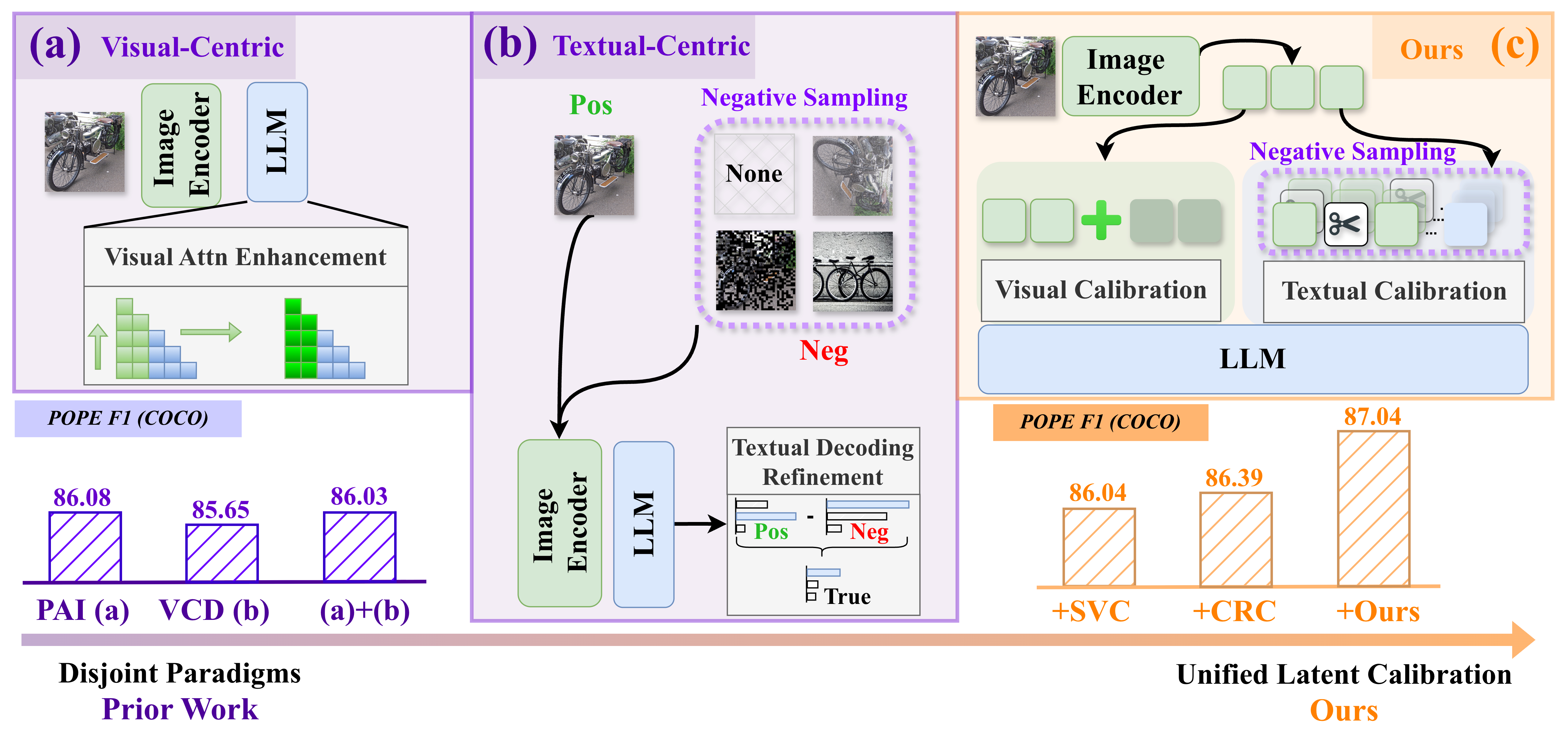} % 使用您的新图 intro.jpg
 \caption{
    \textbf{Disjoint Paradigms vs. Our Unified Latent Calibration.} \textbf{(a)} Prior work includes Visual Attention Enhancement \cite{liu2024paying} and \textbf{(b)} Textual Decoding Refinement at final logits \cite{leng2024mitigating}. Naively combining these disjoint paradigms \textbf{(a)+(b)} degrades performance, highlighting conflicting signals. \textbf{(c)} Our \textbf{Unified Latent Calibration} is a unified system operating entirely at the representation level, using the vision token as a single source for both SVC and CRC, achieving superior and synergistic results.
}
 \label{fig:intro}
\end{figure}
%------------------------------------

%----------- MOTIVATION FIGURE (Stays the same) -----------
\begin{figure*}[t]
 \centering
 \includegraphics[width=1\linewidth]{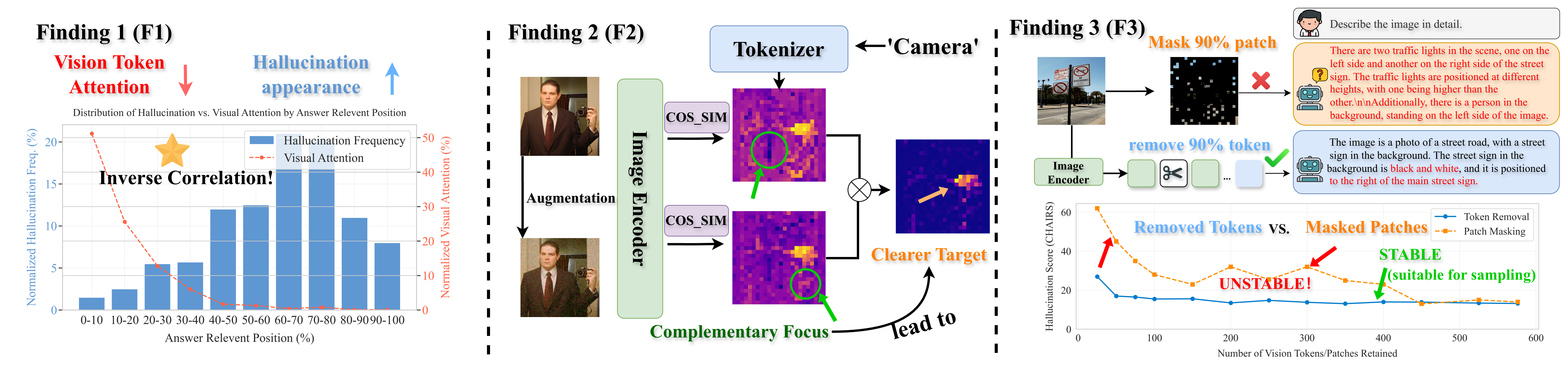}
 \caption{
    \textbf{Our Three Core Findings.}
    \textbf{(F1) Diagnosing the Imbalance: Inverse Correlation.} Visual attention decays sharply as generation proceeds, while hallucination frequency surges where visual grounding is weakest.
    \textbf{(F2) Enabling Enhancement: Semantic Complementarity.} Original and augmented image attentions show complementary focus (e.g., on 'Camera'); their synergy enables enhanced visual grounding.
    \textbf{(F3) Enabling Calibration: Superiority of Information-Gap.} Latent-space token removal (\textit{information-gap}) generates stable, grounded hallucinations, proving more suitable for bias probing than unstable, noisy pixel-level masking (\textit{modality-gap}).
}
 \label{fig:motivation}
\end{figure*}
%-----------------------------------------

Multimodal learning, Large Language Models (LLMs) and Multimodal Large Language Models (MLLMs) represent significant progress in artificial intelligence, showing powerful capabilities in understanding and reasoning about the world \cite{liu2024improved,Qwen-VL,zhu2023minigpt,dai2023instructblip,chen2023shikra,bao2026recent}. The vision token is the key bridge connecting image signals and text signals in these models, making it a central focus of study \cite{neotowards,fu2025hidden,xing2025demystifying}. However, the reliability of MLLMs is undermined by a critical flaw: hallucination \cite{liu2024survey,leng2024mitigating,huang2024opera}, which refers to the fact that MLLMs generate fluent text that contradicts the visual evidence, posing a major barrier to realistic use.

% Introduce Imbalance and F1 as Diagnosis
At its core, as commonly discussed in prior works \cite{liu2024paying,park2025halloc,sarkar2024mitigating},  MLLM hallucination stems from a fundamental \textbf{imbalance}: the visual signal progressively weakens during outputting, while the model's strong internal language prior takes over. Our analysis confirms this imbalance (Figure~\ref{fig:motivation}(\textbf{F1})), showing that visual attention decays sharply as text generation, precisely when hallucinations are most likely to appear, posing a core problem: visual signals gradually weaken over time compared to language signals.

% Introduce Disjointed Solutions and Trade-offs
Existing training-free strategies try to fix this imbalance using disjoint approaches, as depicted in Figure~\ref{fig:intro}. Some focus on \textbf{Visual Attention Enhancement} (Fig.~\ref{fig:intro}(a)) by boosting attention weights to amplify the visual signal \cite{liu2024paying,yin2025clearsight,tang2025seeing}. Others apply \textbf{Textual Decoding Refinement} (Fig.~\ref{fig:intro}(b)), using negative samples for contrastive decoding at the final output logits to suppress text inertia \cite{leng2024mitigating,zhangself,zhao2025cross,fang2025grounding}. But \textbf{\textit{why is unification necessary?}} We contend these separate methods face a critical trade-off:
\begin{itemize}
    \item \textit{Why enhancing vision alone falls short:} Simply boosting the visual signal (e.g., strengthening attention) often isn't enough to ensure accuracy. The language model component has strong ingrained tendencies (text inertia) \cite{lihidden} that can still dominate the text generation, especially as the influence of the image naturally weakens over longer outputs (F1). As confirmed by prior studies \cite{huoself,leng2024mitigating}, this language inertia is a persistent and inevitable challenge.
    \item \textit{Why suppressing language alone has drawbacks:} Methods focusing only on correcting the language model's text inertia often rely on creating negative examples by distorting the input image (\textit{modality-gap}) \cite{leng2024mitigating,chen2025ict}. Our analysis (F3) shows these distorted images create unreliable contents, results are unstable and unrelated to the visual detail, bringing extra noise. Relying on noisy signals for correction means the answers can become unpredictable.
\end{itemize}
Therefore, MLLM hallucination reflects a systemic imbalance requiring a holistic solution.

% Introduce Challenge (with evidence!) and Pivot to Vision Token
However, designing such a unified framework is a non-trivial challenge. A simple, ``patchwork" combination of these disjoint methods is not a principled solution. As we empirically demonstrate in Figure~\ref{fig:intro}, naively combining an attention enhancement method (PAI \cite{liu2024paying}) with a decoding refinement method (VCD \cite{leng2024mitigating}) actually performs no improvement. This happens because the two methods aren’t designed to work together. One boosts visual details from the original image at the attention stage, while the other suppresses signals at the logit level using a negative sample. Furthermore, their calibration timings are different: one intervenes during the layer-level processing, while the other applies a post-correction at the final output stage. 

So \textbf{\textit{How to achieve true, synergistic unification?}} Recognizing this obstacle, we argue a framework must operate cohesively at the same level. We turned our focus towards the nexus of the vision-language interaction: the \textit{vision token}. Could manipulating this core asset directly offer a path to a unified latent representation calibration?

By systematically investigating the vision token's potential, we uncovered other two foundational \textbf{F}indings (Figure~\ref{fig:motivation}) that reveal the key mechanisms for restoring balance and directly enable our unified design:
\begin{itemize}
    \item \textbf{(F2) Semantic Complementarity:} We discovered that \textit{augmented} vision tokens offer complementary semantics, providing a crucial base to build a richer visual anchor for enhancement, as similarly  explored in \cite{gizdov2025seeing,dahou2025vision,xing2025demystifying}.
    \item \textbf{(F3) Information-Gap for Negative Sampling:} We demonstrated that an \textit{information-gap} (by removing tokens in latent space) yields a superior, in-distribution probe for isolating bias, compared to noisy \textit{modality-gap}  based image-level-distorting methods.
\end{itemize}

% Introduce Unified Framework based on F2/F3
These findings reveal that the vision token possesses the potential to serve two distinct roles simultaneously, which corresponds to two \textbf{\textit{fates}}. Based on this, we propose the first unified, training-free framework that operates at the intermediate representation level (Figure~\ref{fig:intro}(c)). It derives all calibration signals from the vision token:
\begin{itemize}
    \item For enhancement, our \textbf{Synergistic Visual Calibration (SVC)} module (based on F2) uses \textit{augmented} tokens to inject a comprehensive visual context, counteracting visual fading (F1).
    \item For suppression, our \textbf{Causal Representation Calibration (CRC)} module (based on F3) uses \textit{pruned} tokens to craft ``latent-space negative samples'' for precisely purifying model internal biases.
\end{itemize}

% Contributions and Conclusion
By addressing both sides of the vision-language imbalance through this principled manipulation of vision tokens, our contributions can be summarized as:
\begin{enumerate}
    \item We reframe hallucination mitigation as a vision-language balance problem, highlighting the limitations of disjointed approaches and demonstrating the failure of naive combination. This reframing is grounded in our systematic analysis of vision tokens, which reveals their dual potential for both enhancement and calibration.
    \item We propose the first \textbf{unified latent calibration framework} that harmonizes enhancement and suppression by leveraging the dual potential of vision tokens, operating entirely on intermediate representations.
    \item We introduce SVC and CRC as novel, efficient modules instantiating this unified method for targeted enhancement and precise suppression.
\end{enumerate}
Our framework demonstrates superior performance and efficiency across diverse benchmarks, validating our unified, vision-token-centric approach.

\section{Methodology}
\label{sec:methodology}
This section details our proposed unified framework for mitigating MLLM hallucinations by restoring the vision-language balance. We first establish the necessary notation in Sec.~\ref{sec:preliminaries}. Then, we present the overall architecture of our unified framework in Sec.~\ref{sec:overview}. Finally, we elaborate on the two core modules: the Synergistic Visual Calibration (SVC) module in Sec.~\ref{sec:svc}, which leverages semantic complementarity (F2) to counteract visual fading (F1), and the Causal Representation Calibration (CRC) module in Sec.~\ref{sec:crc}, which employs the information-gap principle (F3) for precise bias suppression.
%----------- OVERVIEW FIGURE -----------
\begin{figure*}[t]
    \centering
    \includegraphics[width=1\linewidth]{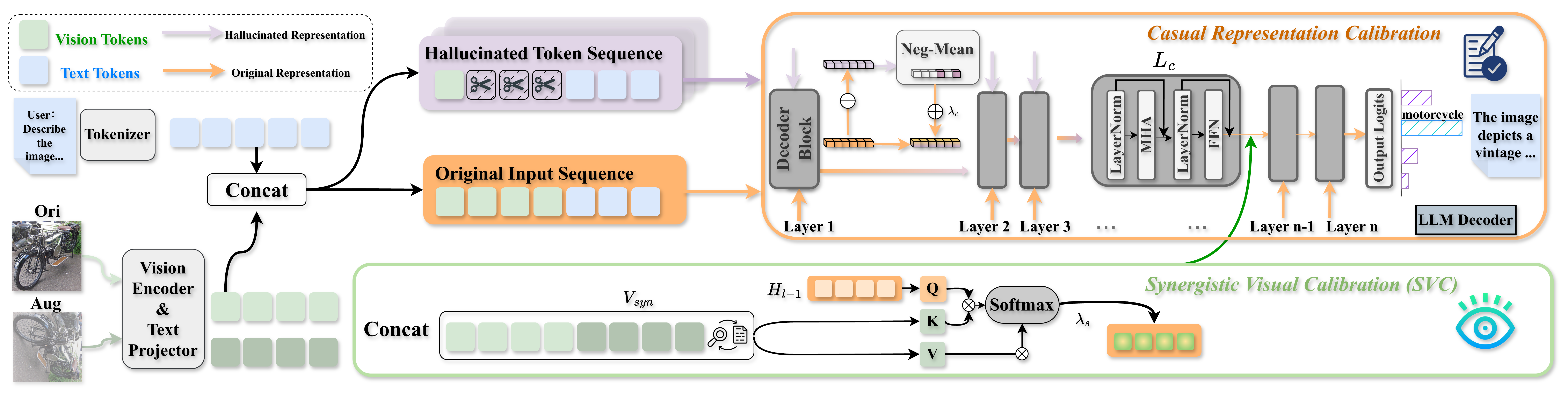}
    \caption{
        \textbf{Overview of our unified framework.} The model processes an original input stream (orange path) and a parallel hallucination-probe stream (purple path) derived from pruned vision tokens. Our \textbf{Synergistic Visual Calibration (SVC)} module injects complementary visual context from augmented images into a critical middle layer ($L_c$) to counteract visual fading. Simultaneously, the \textbf{Causal Representation Calibration (CRC)} module uses the differential representations between the two streams to purify hidden states in shallow layers ($1 \dots L_c$), suppressing linguistic priors.
    }
    \label{fig:overview}
\end{figure*}
%-----------------------------------------

%----------- CRC ILLUSTRATION FIGURE -----------
\begin{figure}[t]
    \centering
    \includegraphics[width=0.9\linewidth]{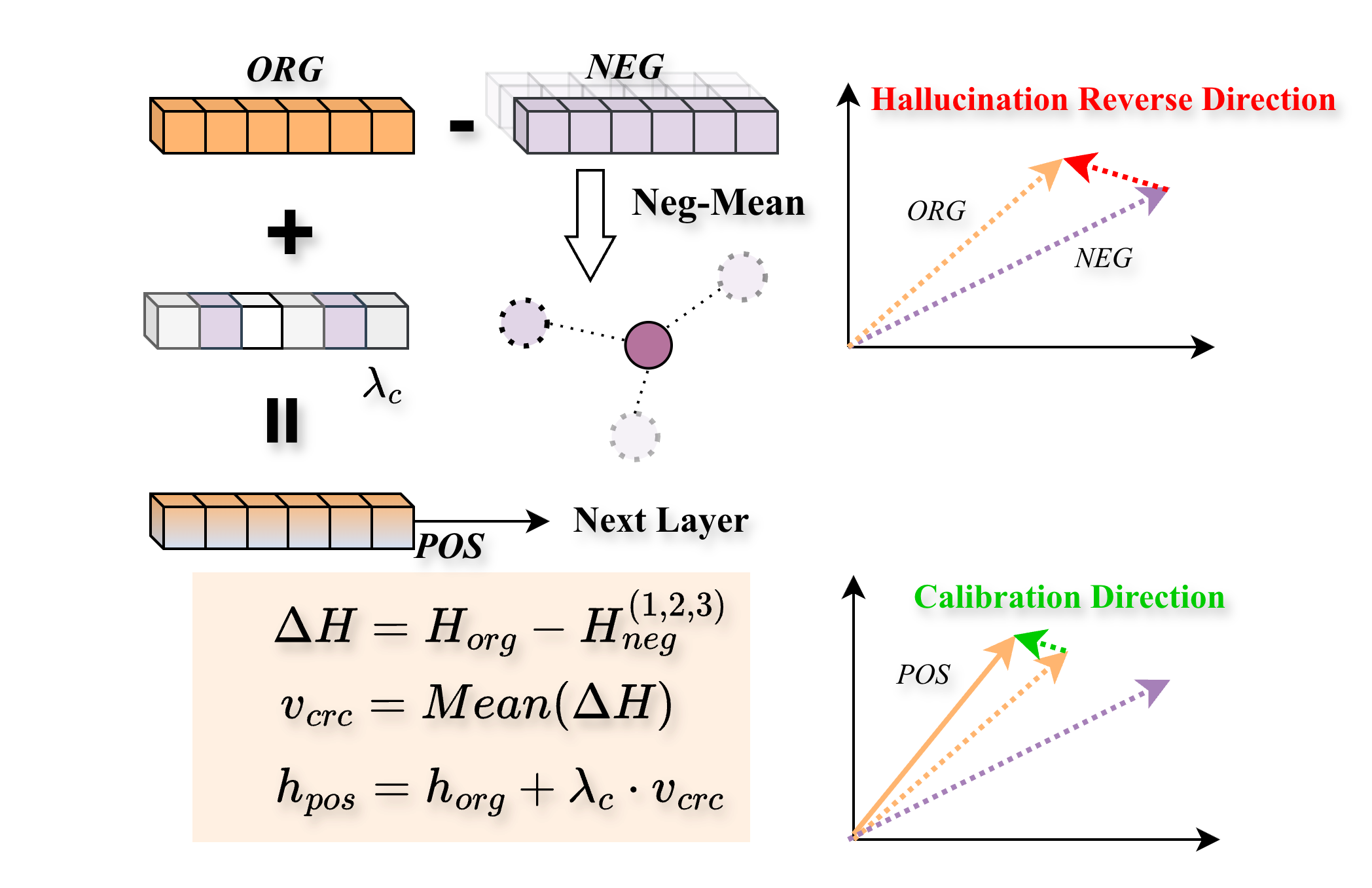}
    \caption{
        \textbf{Illustration of the Causal Representation Calibration (CRC) mechanism.} By subtracting the hallucinated representation ($\mathbf{H}_{\text{neg}}$) from the original ($\mathbf{H}_{\text{org}}$), we obtain a differential vector ($\Delta \mathbf{H}$). Averaging these vectors across multiple negative samples yields a stable hallucination direction ($\mathbf{v}_{\text{crc}}$). The final calibration modifies the original representation away from this direction to produce a purified output ($\mathbf{H}_{\text{pos}}$).
    }
    \label{fig:method}
\end{figure}
%-----------------------------------------
\subsection{Preliminaries}
\label{sec:preliminaries}

We consider a standard MLLM parameterized by $\theta$ that autoregressively generates a response $\mathbf{y} = \{y_1, \ldots, y_{L_y}\}$ given an image $I$ and query $T$. The image is encoded into vision tokens $\mathbf{V} \in \mathbb{R}^{N_v \times d}$, and the query into text tokens $\mathbf{Q} \in \mathbb{R}^{N_q \times d}$. These form the initial context $\mathbf{X}_{\text{context}} = [\mathbf{V}; \mathbf{Q}]$ for the $L$-layer Transformer decoder $\mathcal{D}$.

At generation step $t$, the decoder processes the input embeddings $\mathbf{H}_t^{(0)}$ (derived from $\mathbf{X}_{\text{context}}$ and previous tokens $\mathbf{y}_{<t}$) layer by layer. The output hidden states from any intermediate layer $l \in [1, L]$ are denoted as:
\begin{equation}
    \mathbf{H}_t^{(l)} = \mathcal{D}^{(1 \dots l)}(\mathbf{H}_t^{(0)}),
    \label{eq:intermediate_hidden_states}
\end{equation}
which are crucial for our intervention modules. The probability distribution for the next token $y_t$ is obtained via a softmax over the logits function $f_\theta$:
\begin{equation}
    y_t \sim p_\theta(y_t | I, T, \mathbf{y}_{<t}) = \text{softmax}(f_\theta(I, T, \mathbf{y}_{<t})).
    \label{eq:autoregressive_logits}
\end{equation}
This proceeds iteratively until meeting stopping criterion.

% \subsection{Overview}
% \label{sec:overview}
% Our unified, training-free framework (Figure~\ref{fig:overview}) restores vision-language balance by repurposing vision tokens for two complementary roles. To address \textit{visual fading}, our \textbf{Synergistic Visual Calibration (SVC)} module injects enriched visual context (from original and augmented images) into a critical middle layer via attention. To counteract \textit{text inertia}, our \textbf{Causal Representation Calibration (CRC)} module intervenes in shallow layers: it crafts latent-level negative samples via token pruning to distill a stable hallucination direction vector, which is subtracted from the main stream to purify hidden states. For efficiency, SVC intervenes at a single layer (e.g., layer 16 \cite{wang2025towards,kaduri2025s,chen2025rethinking}), while CRC applies calibration from initial layers to the target layer. All operations act directly on intermediate representations (bypassing decoding) to maximize inference efficiency.

\subsection{Overview}
\label{sec:overview}
Our unified, training-free framework (Figure~\ref{fig:overview}) restores the vision-language balance by repurposing vision tokens for two complementary roles. To address \textit{visual fading}, our \textbf{Synergistic Visual Calibration (SVC)} module injects an enriched visual context (from original and augmented images) into a critical middle layer via attention. To counteract \textit{text inertia}, our \textbf{Causal Representation Calibration (CRC)} module intervenes in shallow layers. It crafts latent-level negative samples via token pruning to distill a stable hallucination direction vector, which is then subtracted from the main computational stream to purify the hidden states. For efficiency, SVC intervenes at a single layer (e.g., layer 16 \cite{wang2025towards,kaduri2025s,chen2025rethinking}), while CRC applies its calibration from the initial layers up to targeted layer. All operations are performed directly on the intermediate representation, bypassing decoding, thereby maximizing inference efficiency.

\subsection{Synergistic Visual Calibration (SVC)}
\label{sec:svc}

To counteract the visual fading phenomenon (F1), we introduce the SVC module, motivated by our finding on semantic complementarity (F2) and prior works \cite{xing2025demystifying,gizdov2025seeing,dahou2025vision,zoulook}.

\paragraph{Synergistic Visual Context Construction.}
Given an image $I$, we create an augmented version $I_{\text{aug}}$
  by applying random horizontal flipping, Gaussian blur with radius 5, and salt-and-pepper noise (intensity = 0.2). Both are processed to obtain their vision token, $\mathbf{V}, \mathbf{V}_{\text{aug}} \in \mathbb{R}^{N_v \times d}$. We concatenate them to form a synergistic visual memory bank:
\begin{equation}
    \mathbf{V}_{\text{syn}} = [\mathbf{V}; \mathbf{V}_{\text{aug}}] \in \mathbb{R}^{2N_v \times d}.
    \label{eq:svc_concat}
\end{equation}

\paragraph{Parameter-Free Visual Injection.}
SVC intervenes at a single, pre-defined middle layer $L_c$. At generation step $t$, we intercept the hidden state sequence $\mathbf{H}_t^{(L_c-1)}$ from the preceding layer, which serves as the \textbf{Query}. The set $\mathbf{V}_{\text{syn}}$ serves as both \textbf{Key} and \textbf{Value}. A visual context sequence $\mathbf{C}_t$ is computed via scaled dot-product attention:
\begin{equation}
    \mathbf{C}_t = \text{softmax}\left(\frac{\mathbf{H}_t^{(L_c-1)} (\mathbf{V}_{\text{syn}})^T}{\sqrt{d}}\right) \mathbf{V}_{\text{syn}}.
    \label{eq:svc_attention}
\end{equation}
As implemented, the resulting context $\mathbf{C}_{t}$ is integrated via interpolation, blending it with the original hidden state:
\begin{equation}
    \mathbf{H}'^{(L_c)}_t = (1 - \lambda_s) \cdot \mathbf{H}^{(L_c)}_{t} + \lambda_s \cdot \mathbf{C}_{t},
    \label{eq:svc_update}
\end{equation}
where $\lambda_s$ is a hyperparameter controlling the interpolation ratio. This updated hidden state $\mathbf{H}'^{(L_c)}_t$ proceeds to the next layer, re-engaging the model with a richer visual context.
\subsection{Causal Representation Calibration (CRC)}
\label{sec:crc}

Distinct from prior methods operating on output logits, CRC intervenes on intermediate representations to neutralize the ``hallucination direction'', visualized in Figure~\ref{fig:method}.

\paragraph{Probing for the Hallucination Direction.}
To isolate intrinsic biases, We craft ``latent space negative samples'' based on our information-gap principle (F3). Given the original vision token sequence $\mathbf{V} \in \mathbb{R}^{N_v \times d}$, we generate $K$ negative samples, $\{\mathbf{V}_{\text{neg}}^{(k)}\}_{k=1}^K$, by randomly pruning $\mathbf{V}$ to retain only $N_h = 5$ tokens.

At the first step $t = 0$, we run parallel forward passes through the layers for the original and negative samples:
\begin{align}
    \mathbf{H}_{0, \text{org}}^{(l)} &= \mathcal{D}^{(1 \dots l)}(\,[\mathbf{V}; \mathbf{Q}]\,), \\
    \mathbf{H}_{0, \text{neg}}^{(l, k)} &= \mathcal{D}^{(1 \dots l)}(\,[\mathbf{V}_{\text{neg}}^{(k)}; \mathbf{Q}]\,).
\end{align}
The difference, $\Delta \mathbf{H}^{(l,k)} = \mathbf{H}_{0, \text{org}}^{(l)} - \mathbf{H}_{0, \text{neg}}^{(l, k)}$, captures the representational shift. We compute the stable hallucination vector $\mathbf{v}_{\text{crc}}^{(l)}$ by averaging this difference:
\begin{equation}
    \mathbf{v}_{\text{crc}}^{(l)} = \frac{1}{K} \sum_{k=1}^{K} \Delta \mathbf{H}^{(l,k)}.
    \label{eq:crc_vector}
\end{equation}
This vector $\mathbf{v}_{\text{crc}}^{(l)}$ acts as our representation probe cache and is re-used for all subsequent processing steps.
\paragraph{Causal Calibration.}
At each generation step $t > 0$, we perform calibration by adjusting the current hidden state. This operation is performed in normalized space to preserve representation stability. First, the original state and the calibration vector are normalized:
\begin{equation}
    \mathbf{h}_{\text{norm}} = \frac{\mathbf{H}_{t, \text{org}}^{(l)}}{||\mathbf{H}_{t, \text{org}}^{(l)}||_2}, \quad \mathbf{v}_{\text{norm}} = \frac{\mathbf{v}_{\text{crc}}^{(l)}}{||\mathbf{v}_{\text{crc}}^{(l)}||_2}.
\end{equation}
Next, a new unnormalized vector $\mathbf{h}_{\text{crc}}$ is computed by linearly combining the normalized vectors:
\begin{equation}
    \mathbf{h}_{\text{crc}} = \mathbf{h}_{\text{norm}} + \lambda_c \cdot \mathbf{v}_{\text{norm}}.
    \label{eq:12}
\end{equation}
Finally, the corrected hidden state $\mathbf{H}_{t, \text{pos}}^{(l)}$ is obtained by re-normalizing this vector and scaling it back to the original magnitude of $\mathbf{H}_{t, \text{org}}^{(l)}$:
\begin{equation}
    \mathbf{H}_{t, \text{pos}}^{(l)} = \frac{\mathbf{h}_{\text{crc}}}{||\mathbf{h}_{\text{crc}}||_2} \cdot ||\mathbf{H}_{t, \text{org}}^{(l)}||_2,
    \label{eq:crc_calibration}
\end{equation}
where $\lambda_c$ is a hyperparameter. This calibrated representation proceeds to the next layer. This process is repeated for all layers $l=1, \dots, L_c$.
%----------- SCM FIGURE -----------
\begin{figure}[t]
\centering
\includegraphics[width=0.8\linewidth]{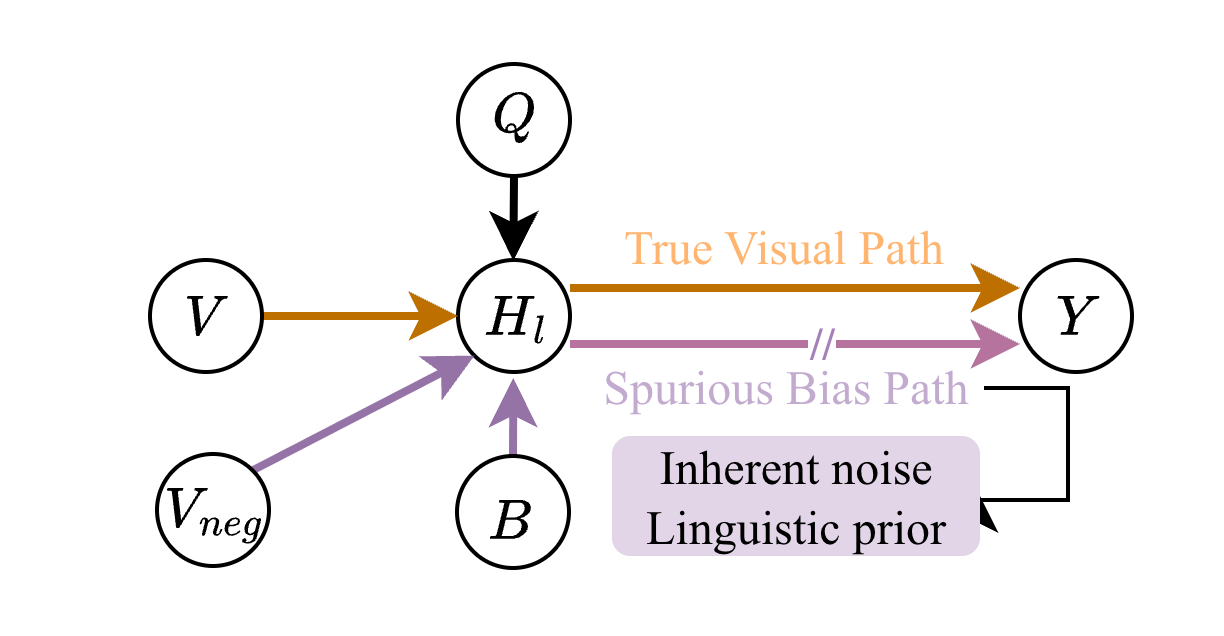}
\caption{
    \textbf{The simplified Structural Causal Model (SCM) for MLLM hallucination.} We posit that hallucination arises from a spurious causal path from the model's intrinsic bias ($B$) to its latent representation ($H^{(l)}$), which confounds the true visual path from the image ($V$).
}
\label{fig:theory}
\end{figure}
%-----------------------------------------

% \paragraph{Theoretical Justification.}
% Our approach is formally justified by Structural Causal Models (SCM). As shown in Figure~\ref{fig:theory}, we model hallucination as arising from a spurious causal path from intrinsic bias ($B$) to the latent representation ($H_t^{(l)}$). Our essential negative sample acts as a causal intervention, isolating the influence of this path. The differential vector $\mathbf{v}_{\text{crc}}$ thus estimates the pure causal effect of vision, disentangled from bias, including MLLM inherent noise during training and linguistic prior caused by its projector-based multimodal alignment architecture \cite{alayrac2022flamingo,liu2024improved,Qwen-VL}. The final calibration (Eq.~\ref{eq:crc_calibration}) is a counterfactual adjustment that steers the representation towards a more faithful, visually-grounded state. A full theoretical derivation is provided in the Appendix.
\paragraph{Theoretical Justification.}
Our Causal Representation Calibration (CRC) is grounded in Structural Causal Models (SCM). We model hallucination (Figure~\ref{fig:theory}) as arising from a spurious causal path where intrinsic bias ($B$) confounds the true visual path ($V \to H_t^{(l)}$). This bias includes inherent noise and linguistic priors. Our essential negative sample, derived from a degraded visual input $V_{\text{neg}}$, acts as an \textit{in-distribution} causal probe by retaining the original image's structural properties.

The core theoretical insight is that our differential vector $\mathbf{v}_{\text{crc}}$ isolates the causal effect of the \textit{missing visual information}. Following a commonly adopted local linear approximation supported by \cite{huang2025deciphering,nguyen2025grains,gan2025textual,zhoumitigating}, we estimate the hidden states decompose as:
\begin{align}
    \mathbf{H}_{t, \text{org}}^{(l)} &\approx \mathcal{E}(V) + \mathcal{E}_{\text{shared}}(Q, B) \\
    \mathbf{H}_{t, \text{neg}}^{(l)} &\approx \mathcal{E}(V_{\text{neg}}) + \mathcal{E}_{\text{shared}}(Q, B)
\end{align}
where $\mathcal{E}(\cdot)$ is the causal effect function. Subtracting these cancels the shared query ($Q$) and bias ($B$) effects. Thus, our representation probe captures the pure visual difference:
\begin{equation}
    \mathbf{v}_{\text{crc}}^{(l)} \approx \mathcal{E}(V - V_{\text{neg}}),
    \label{eq:appendix_causal_core_interpret}
\end{equation}
which represents the pure signal lost due to degradation. The calibration step Eq.(\ref{eq:12}) uses this to perform a counterfactual adjustment, towards visual truth representation. 

\textbf{Summary.} \quad Eq.(\ref{eq:appendix_causal_core_interpret}) provides a principled way to estimate the pure visual signal to cancel shared effects of the text prior and internal bias. 
\section{Experiments}
\begin{table*}[t]
\centering
% 使用resizebox控制表格宽度为\textwidth（文本宽度）
\renewcommand{\arraystretch}{0.9} % 减小行高

\resizebox{\textwidth}{!}{
\begin{tabular}{llcccccccc}
\toprule
\multirow{2}{*}{Benchmark} & \multirow{2}{*}{Method} & \multicolumn{2}{c}{LLaVA-1.5} & \multicolumn{2}{c}{MiniGPT-4} & \multicolumn{2}{c}{Shikra} & \multicolumn{2}{c}{InstructBLIP} \\
\cmidrule(lr){3-4} \cmidrule(lr){5-6} \cmidrule(lr){7-8} \cmidrule(lr){9-10}
& & Avg. Accuracy $\uparrow$ & Avg. F1 $\uparrow$ & Avg. Accuracy $\uparrow$ & Avg. F1 $\uparrow$ & Avg. Accuracy $\uparrow$ & Avg. F1 $\uparrow$ & Avg. Accuracy $\uparrow$ & Avg. F1 $\uparrow$ \\
\midrule
\multirow{6}{*}{MSCOCO} 
& Vanilla & 84.79 & 85.61 & 76.76 & 76.82 & 81.32 & 82.01 & 84.36 & 84.64 \\
& VCD \scriptsize{\textcolor{gray}{\texttt{[CVPR'24]}}} & 84.80 & 85.65 & 76.01 & 76.28 & 81.34 & 82.25 & 84.81 & \textbf{85.28} \\
& PAI \scriptsize{\textcolor{gray}{\texttt{[ECCV'24]}}} & 85.85 & 86.08 & 75.64 & \underline{77.57} & 81.30 & 80.81 & 83.98 & 84.02 \\
& VISTA \scriptsize{\textcolor{gray}{\texttt{[ICML'25]}}} & \underline{86.15} & \underline{86.29} & 76.06 & 76.80 & 82.44 & 82.47 & \underline{84.87} & 84.95 \\
& ONLY \scriptsize{\textcolor{gray}{\texttt{[ICCV'25]}}} & 86.03 & 86.22 & \textbf{76.98} & \textbf{77.62} & \underline{82.75} & \underline{82.85} & \textbf{84.90} & \underline{85.01} \\
\rowcolor{highlightcolor}
& Ours & \textbf{86.79} & \textbf{87.04} & \underline{76.78} & 77.29 & \textbf{83.84} & \textbf{83.58} & 84.59 & 84.97 \\
\midrule
\multirow{6}{*}{AOKVQA} 
& Vanilla & 77.23 & 80.62 & 72.80 & 72.99 & 78.67 & 80.03 & 78.92 & 81.51 \\
& VCD \scriptsize{\textcolor{gray}{\texttt{[CVPR'24]}}}& 76.29 & 79.99 & 71.27 & 71.93 & 78.52 & 80.24 & 78.95 & 81.52 \\
& PAI \scriptsize{\textcolor{gray}{\texttt{[ECCV'24]}}}  & 78.65 & 79.82 & \textbf{73.20} & \underline{73.22}  & 79.97 & 81.07 & 79.04 & 81.77 \\
& VISTA \scriptsize{\textcolor{gray}{\texttt{[ICML'25]}}} & \underline{81.23} & 81.65 & 71.82 & 72.34 & 80.12 & 81.24 & 80.09 & 82.30 \\ 
& ONLY \scriptsize{\textcolor{gray}{\texttt{[ICCV'25]}}} & 80.55 & \underline{82.24} & 72.54 & 72.56 & \underline{81.64} &\underline{81.78}  & \textbf{80.76} & \textbf{82.58} \\
\rowcolor{highlightcolor}
& Ours & \textbf{82.23} & \textbf{83.82} & \underline{73.13} & \textbf{73.69} & \textbf{81.87} & \textbf{82.18} & \underline{80.24} &  \underline{82.48}\\
\midrule
\multirow{6}{*}{GQA} 
& Vanilla & 78.76 & 80.79 & 70.81 & 73.08 & 78.59 & 79.70 & 77.31 & 79.08 \\
& VCD \scriptsize{\textcolor{gray}{\texttt{[CVPR'24]}}} & 79.36 & 80.92 & 70.60 & 72.98 & 78.89 & 79.96 & 77.11 & 78.98 \\
& PAI \scriptsize{\textcolor{gray}{\texttt{[ECCV'24]}}}  & 79.80 & 81.12 & 71.12 & 73.20 & 79.23 & 80.16 & 77.56 & 79.18 \\
& VISTA \scriptsize{\textcolor{gray}{\texttt{[ICML'25]}}} & \underline{80.89} & \underline{82.37} & 71.13 & 73.19 & 79.59 & 80.20 & \underline{78.01} & \textbf{80.34} \\
& ONLY \scriptsize{\textcolor{gray}{\texttt{[ICCV'25]}}} & 80.44 & 82.17 & \textbf{71.99} & \textbf{73.89} & \underline{80.28} & \underline{81.02} & 77.98 & \underline{80.32} \\
\rowcolor{highlightcolor}
& Ours & \textbf{81.54} & \textbf{83.38} & \underline{71.40} & \underline{73.37} & \textbf{81.24} & \textbf{81.44} & \textbf{78.11} & 80.29 \\
\bottomrule
\end{tabular}
}
\caption{Evaluation results on POPE benchmark across four MLLMs. Results show averaged accuracy and F1 scores in \% computed across \textbf{random}, \textbf{popular}, and \textbf{adversarial} object splits. Best and second best results are bolded and underlined, respectively.}
\label{tab:pope_eval}
\end{table*}

\begin{table*}[t]
\centering
\renewcommand{\arraystretch}{0.9} % 减小行高
\resizebox{\textwidth}{!}{%
\begin{tabular}{llcccccccc}
\toprule
\multirow{2}{*}{Max Tokens} & \multirow{2}{*}{Method} & \multicolumn{2}{c}{LLaVA-1.5 } & \multicolumn{2}{c}{MiniGPT-4} & \multicolumn{2}{c}{Shikra} & \multicolumn{2}{c}{InstructBLIP} \\
\cmidrule(lr){3-4} \cmidrule(lr){5-6} \cmidrule(lr){7-8} \cmidrule(lr){9-10}
& & $\text{CHAIR}_\text{Sen} \downarrow$ & $\text{CHAIR}_\text{Ins} \downarrow$ & $\text{CHAIR}_\text{Sen} \downarrow$ & $\text{CHAIR}_\text{Ins} \downarrow$ & $\text{CHAIR}_\text{Sen} \downarrow$ & $\text{CHAIR}_\text{Ins} \downarrow$ & $\text{CHAIR}_\text{Sen} \downarrow$ & $\text{CHAIR}_\text{Ins} \downarrow$ \\
\midrule
\multirow{5}{*}{64 tokens} 
& Vanilla & 25.4 & 8.9 & 25.0 & 9.4 & 25.4 & 8.8 & 26.8 & 9.2 \\
& VCD \scriptsize{\textcolor{gray}{\texttt{[CVPR'24]}}} & 24.1 & 7.9 & 25.4 & 9.4 & 23.2 & 8.1 & 26.8 & 9.3 \\
& VISTA \scriptsize{\textcolor{gray}{\texttt{[ICML'25]}}} & 21.4 &\underline{8.1} & \textbf{22.2} & \textbf{8.2} & \underline{19.5} & \underline{6.7} & \underline{24.4} & \underline{8.8} \\
& ONLY \scriptsize{\textcolor{gray}{\texttt{[ICCV'25]}}}& \underline{19.2} & \textbf{7.5} & 24.5 & 9.3 & 21.4 & 7.5 &  25.7 & 9.0 \\
\rowcolor{highlightcolor}
& Ours & \textbf{18.1} & 8.4 & \underline{23.8} & \underline{8.3} & \textbf{16.7} & \textbf{5.9} & \textbf{24.2} & \textbf{7.9} \\
\midrule
\multirow{5}{*}{128 tokens} 
& Vanilla & 48.9 & 14.9 & 36.4 & 11.3 & 56.8 & 15.6 & 43.2 & 12.4  \\
& VCD \scriptsize{\textcolor{gray}{\texttt{[CVPR'24]}}} & 47.4 & 12.9 & 35.4 & 10.8 & 49.4 & 13.4 & 42.1 & 12.3 \\
& VISTA \scriptsize{\textcolor{gray}{\texttt{[ICML'25]}}} & \underline{38.6} & \underline{11.7} & \underline{31.2} & \underline{10.2} & \underline{36.4} & \underline{10.8} & \underline{36.2} & \textbf{10.4} \\
& ONLY \scriptsize{\textcolor{gray}{\texttt{[ICCV'25]}}} & 42.2 & 12.4 & 33.9 & 10.6 & 38.2 & 11.4 & 40.2 & 11.6 \\
\rowcolor{highlightcolor}
& Ours & \textbf{36.8} & \textbf{11.7} & \textbf{30.6} & \textbf{9.9} & \textbf{33.4} & \textbf{9.9} & \textbf{39.4} & \underline{11.5} \\
\bottomrule
\end{tabular}%
}
\caption{CHAIR hallucination evaluation results. We compare our method to state-of-the-art training-free methods. Maximum new token is set to 64 and 128. Best and second best results are bolded and underlined, respectively.}
\label{tab:chair_eval}
\end{table*}
\begin{figure*}[t]
  \centering
  \includegraphics[width=1\linewidth]{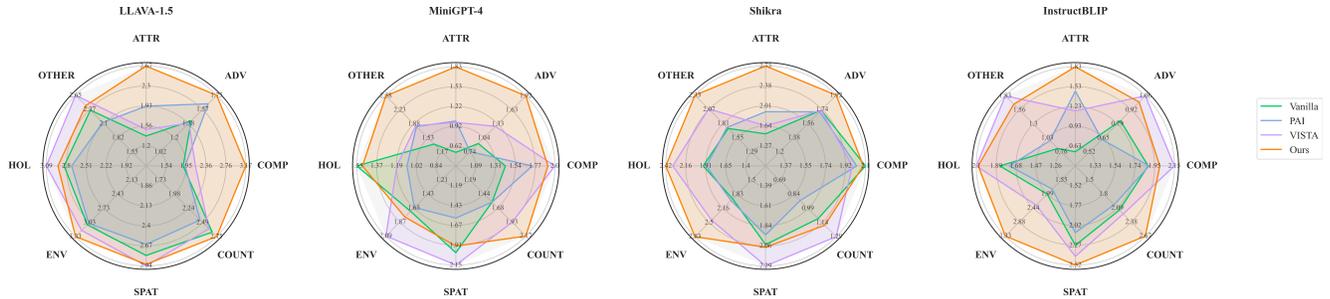} % Assuming you will name your MMHal results radar plot this
  \caption{
      MMHal-Bench Evaluation across four MLLMs. Our method (Ours, orange line) consistently achieves superior performance across all eight categories, demonstrated by the larger area covered in the radar plots compared to Vanilla, PAI, and VISTA.
  }
  \label{fig:mmhal_results}
\end{figure*}

In this section, we validate our framework across multiple MLLM architectures and four popular benchmarks. We first present the experimental configuration (Sec.~\ref{sec:exp_setup}), followed by an extensive evaluation on hallucination (Sec.~\ref{sec:exp_hallucination}) and general comprehensive benchmarks (Sec.~\ref{sec:exp_comprehensive}), and finally we quantify the computational cost (Sec.~\ref{sec:exp_overhead}). 

\subsection{Experimental Setup}
\label{sec:exp_setup}

\paragraph{Model Architectures.}
We mainly evaluate our unified framework on four representative MLLMs with distinct architectural designs include: \textbf{LLaVA-1.5} \cite{liu2024improved} and \textbf{Shikra} \cite{chen2023shikra}, which employ linear projections for visual-textual alignment, and \textbf{MiniGPT-4} \cite{zhu2023minigpt} and \textbf{InstructBLIP} \cite{dai2023instructblip}, which utilize a Q-Former for cross-modal interaction.

\paragraph{Implementation Details.}
We employ the following configuration across all experiments unless stated otherwise. For our Synergistic Visual Calibration (SVC) module, the intervention layer is set to $L_c=16$ for all models, and the scaling factor $\lambda_s$ is set to 0.06. For our Causal Representation Calibration (CRC) module, the pruning retains $N_h=5$ vision tokens, the number of negative samples is $K=3$ for balance of effectiveness and efficiency, and the calibration strength $\lambda_c$ is set to 0.1 by default. 

\subsection{Results on Object Hallucination Benchmarks}
\label{sec:exp_hallucination}

We first evaluate our framework on the two widely adopted benchmarks that assess object hallucination: POPE \cite{li2023evaluating} and CHAIR \cite{rohrbach2018object}. We compare our method against several recent strong training-free based baselines, including VCD \cite{leng2024mitigating}, PAI \cite{liu2024paying}, VISTA \cite{lihidden}, ONLY \cite{wan2025only}.

\noindent \textbf{POPE Evaluation.}  
The POPE benchmark~\cite{li2023evaluating} assesses object hallucination via yes/no questions (e.g., “Is there a [object] in the image?”) across three splits of increasing difficulty. As shown in Table~\ref{tab:pope_eval}, our method consistently outperforms prior work: on the challenging GQA split, it achieves \textbf{81.54\%} accuracy with LLaVA-1.5 and \textbf{78.11\%} with InstructBLIP, demonstrating strong generalizability across architectures (linear projector vs. Q-Former) and datasets (COCO~\cite{lin2014microsoft}, AOKVQA~\cite{schwenk2022okvqa}, GQA~\cite{hudson2019gqa}).

\noindent \textbf{CHAIR Evaluation.}  
CHAIR~\cite{rohrbach2018object} measures hallucinated objects in open-ended captions via instance-level ($\text{CHAIR}_I$) and sentence-level ($\text{CHAIR}_S$) scores (lower is better). Table~\ref{tab:chair_eval} shows our approach achieves the best $\text{CHAIR}_I$ scores: \textbf{18.1} (LLaVA-1.5) and \textbf{16.7} (Shikra) at 64 tokens, and \textbf{30.6} (MiniGPT-4) and \textbf{33.4} (Shikra) at 128 tokens, confirming that our latent-level calibration effectively suppresses ungrounded object generation.

%----------- CHAIR RESULTS TABLE (Placeholder) -----------
\begin{table}[t]
\centering

\resizebox{\linewidth}{!}{%
\begin{tabular}{lcccc}
\toprule
\multirow{2}{*}{Method} & \multicolumn{2}{c}{LLaVA-1.5} & \multicolumn{2}{c}{InstructBlip}  \\

& $\text{Perception}_\text{sum}$ & $\text{Cognition}_\text{sum}$  & $\text{Perception}_\text{sum}$  &$\text{Cognition}_\text{sum}$  \\
\midrule
Vanilla & 1444.23 & 323.14 & 1181.15 & 233.71 \\
VCD & 1445.13 & 324.25 & 1182.34 & 234.95\\
VISTA & 1450.54 & 329.08 & 1189.23 & 236.25  \\
\midrule
\rowcolor{highlightcolor}
Ours & \textbf{1456.28} & \textbf{332.86} & \textbf{1192.76} & \textbf{240.83}  \\
\bottomrule
\end{tabular}%
}
\caption{Evaluation on MME benchmark (Perception and Cognition scores). Our method (Ours) consistently achieves state-of-the-art performance across all tested models, demonstrating robust general capabilities.}
\label{tab:comprehensive_results} % Ensure this label matches your table
\end{table}
%-------------------------------------------------------

\begin{table}[t]
\centering

\resizebox{\linewidth}{!}{
\begin{tabular}{lccc}
\toprule
Method & \makecell{Latency (ms/token)} & \makecell{Throughput (token/ms)} & \makecell{Memory Cost (MB)} \\
\midrule
Greedy & 30.3 (×1.00) & 0.033 (×1.00) & 14257 \\
\hdashline

ICD & 33.03 (×1.09) & 0.030 (×0.91) & 14263  \\
VCD & 72.72 (×2.4) & 0.014 (×0.42) & 14984  \\
VISTA & 33.35 (×1.1) & 0.030 (×0.91) &  15024 \\
\rowcolor{highlightcolor}
Ours & \textbf{32.1 (×1.06)} & \textbf{0.031 (×0.94)} &  14924 \\
\bottomrule
\end{tabular}
}
\caption{Comparison of overhead among different methods (with best results bolded). Latency is measured in milliseconds per token. Throughput is calculated as tokens per millisecond. Memory cost is the peak GPU memory usage in megabytes.}
\label{tab:overhead_comparison_bold}
\end{table}
\subsection{Results on Comprehensive Benchmarks}
\label{sec:exp_comprehensive}

A critical aspect of our framework is its ability to suppress hallucinations without harming the model's general perception and reasoning abilities. We validate this on MMHal-Bench and MME.

\noindent \textbf{MMHal-Bench Evaluation.}
MMHal-Bench \cite{sun2023aligning} assesses hallucination via 96 image-question pairs across eight categories (e.g., colors, counting), focusing on complex visual reasoning. Responses are scored by GPT-4 \cite{achiam2023gpt}. As shown in Figure~\ref{fig:mmhal_results}, our method outperforms Vanilla, PAI \cite{liu2024paying}, and VISTA \cite{lihidden} across all categories on multiple MLLMs, with notable gains in ATTR, and ENV, indicating better use of true visual signals.

\noindent \textbf{MME Evaluation.}
On MME \cite{yin2024survey}, which evaluates 14 perception and cognition abilities, our framework improves general performance and hallucination mitigation. On LLaVA-1.5, we achieve Perception: \textbf{1456.28} and Cognition: \textbf{332.86}, surpassing Vanilla, VCD, and VISTA; similar gains are seen on InstructBLIP (Perception: \textbf{1192.76}, Cognition: \textbf{240.83}). confirming our method enhances MLLM capabilities by better balancing vision and language.

\subsection{Computational Overhead}
\label{sec:exp_overhead}

Our framework is highly efficient, which is highly crucial for a training-free style method. As shown in Table~\ref{tab:overhead_comparison_bold}, it incurs only a 1.06× latency increase over Greedy and faster than VISTA (33.35 ms/token) and VCD (72.72 ms/token). It also uses less peak GPU memory (14,924 MB) than VISTA (15,024 MB) and VCD (14,984 MB). Thus, this proves our method achieves strong hallucination mitigation with minimal computational overhead.

\section{Discussions}
\label{sec:discussion}

We now analyze our framework's components and design choices through ablation studies and visualizations. All discussions are based on experiments conducted on LLaVA-1.5 using POPE benchmarks unless otherwise specified.

\textbf{4.1 Ablation Study}

\begin{table}[t]
\centering

\resizebox{\linewidth}{!}{%
\begin{tabular}{lcccccc}
\toprule
\multirow{2}{*}{Method Configuration} & \multicolumn{2}{c}{COCO} & \multicolumn{2}{c}{AOKVQA}\\
& Acc $\uparrow$ & F1 $\uparrow$ & Acc $\uparrow$ &  F1 $\uparrow$  \\
\midrule
Vanilla LLaVA-1.5 & 84.79 & 85.61 & 77.23 & 80.62  \\
\midrule
+ SVC ($V_{\text{ori}}$ only) & 85.04 & 85.68 & 79.03 & 80.96 \\
+ SVC ($V_{\text{syn}}$, Ours) & 85.55 & 86.04 & 79.43 & 81.73 \\
\midrule
+ CRC (Masked Image neg.) & 84.77 & 85.72 & 79.15 & 81.34 \\
+ CRC (Pruned Token neg., Ours) & 86.11 & 86.39 & 81.65 & 81.98\\
\midrule
\rowcolor{highlightcolor} % Define \highlightcolor in your preamble
Ours (SVC + CRC) & \textbf{86.79} & \textbf{87.04} & \textbf{82.23} & \textbf{83.82} \\
\bottomrule
\end{tabular}%
}
\caption{Ablation study on LLaVA-1.5 (POPE benchmark in \%) evaluating our SVC variants (visual context) and CRC variants (negative sampling strategy). Best results are \textbf{bolded}.}
\label{tab:ablation_study}
\end{table}

\noindent Table~\ref{tab:ablation_study} shows the results of our ablation study. Both our key modules, SVC (using $V_{\text{syn}}$) and CRC (using pruned tokens), improve performance individually over the Vanilla baseline. Notably, the full model integrating both SVC and CRC achieves the best scores across all metrics. This confirms that our two modules are effective and work synergistically, validating our unified design. 

\textbf{4.2 Why SVC works}

\begin{figure}[t]
    \centering
    \includegraphics[width=1\linewidth]{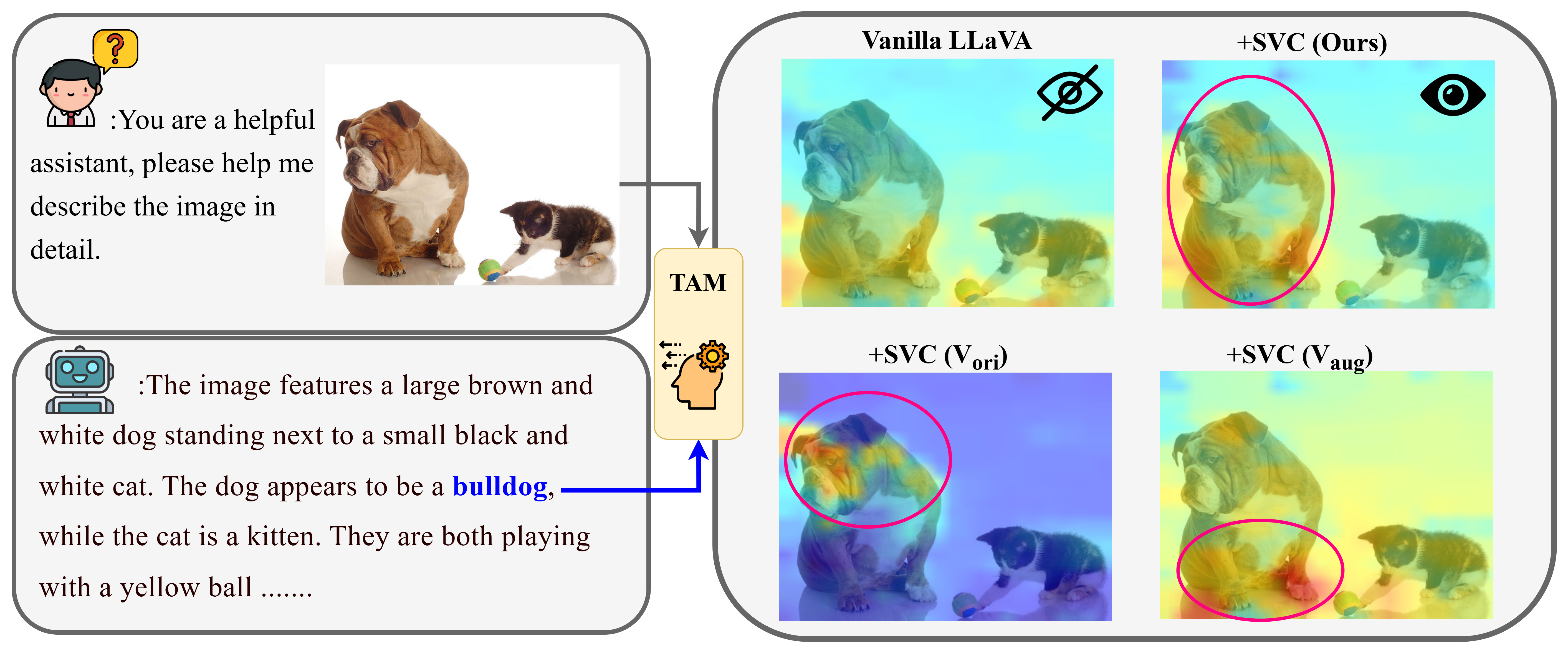} % Assumes file is named SVC.png and in fig/ subdirectory
    \caption{
        \textbf{Visualizing SVC's effect with TAM \cite{li2025token}.} For the token 'bulldog', Vanilla LLaVA shows diffuse attention. $V_{\text{ori}}$ and $V_{\text{aug}}$ alone show different, complementary focus. ($+SVC (\text{Ours})$) integrates these cues, creating a accurate focus on the bulldog.
    }
    \label{fig:svc_tam}
\end{figure}

\noindent To understand SVC, we use Token Activation Mapping (TAM)~\cite{li2025token} to visualize where the model attends (Figure~\ref{fig:svc_tam}). Focusing on the token `bulldog' in a case with unchanged output, we find the Vanilla model exhibits diffuse attention. Using only the original ($V_{\text{ori}}$) or augmented ($V_{\text{aug}}$) image yields distinct, complementary attention patterns (F2). Our full SVC, which fuses both, produces a sharper, more focused map on the bulldog, demonstrating that our SVC leverages complementary visual cues to better ground attention on original image-relevant details.

\textbf{4.3 Why CRC works}

\begin{figure}[t]
    \centering
    \includegraphics[width=0.9\linewidth]{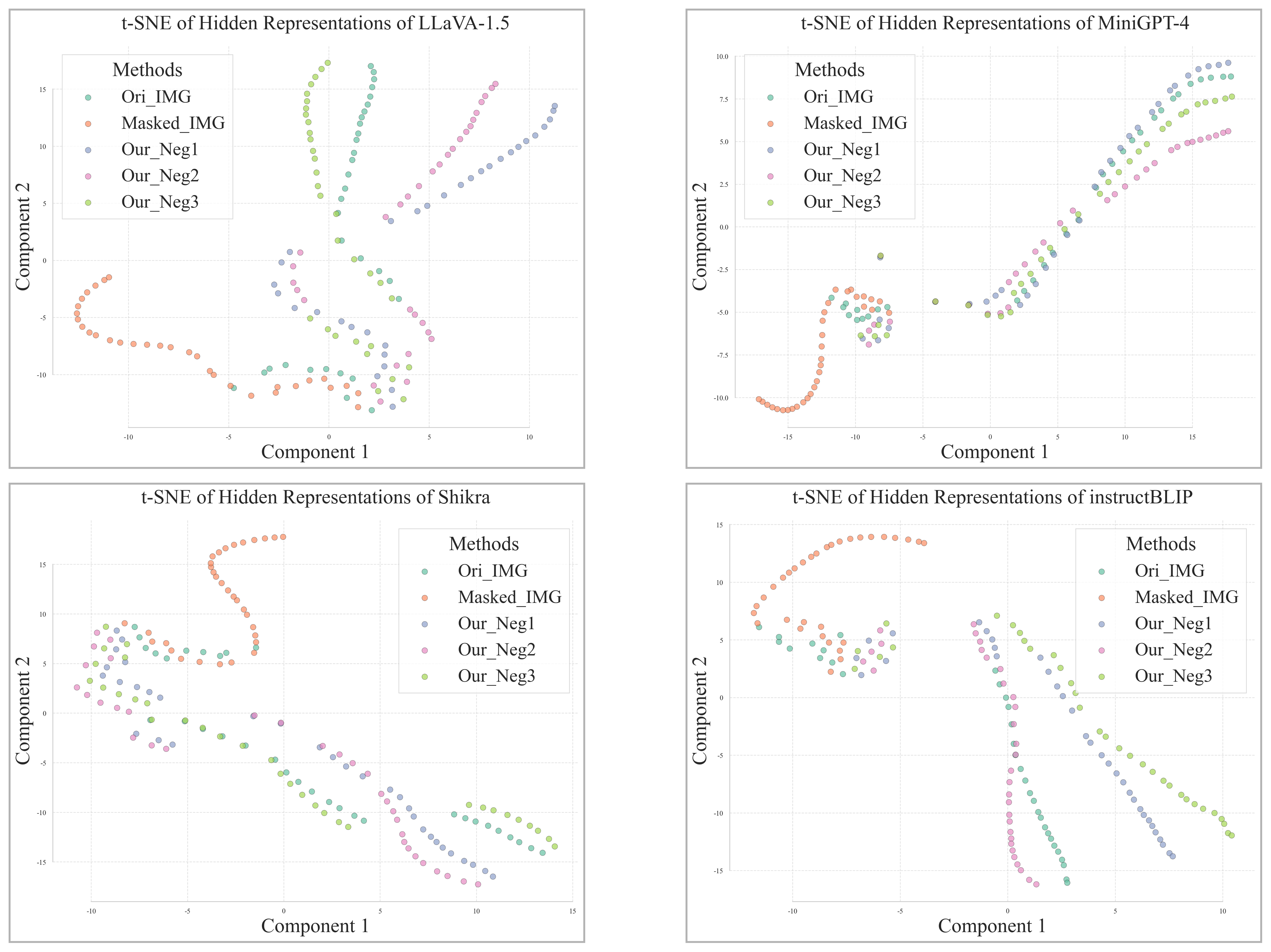} % Assumes file is named CRC.png and in fig/ subdirectory
    \caption{
        \textbf{t-SNE visualization of hidden states.} Our pruned-token negative samples (Our\_Neg1/2/3) consistently cluster near the original (Ori\_IMG) representations, suggesting an \textit{in-distribution} probe. In contrast, Masked\_IMG representations are divergent, indicating a noisy, \textit{out-of-distribution} perturbation.
    }
    \label{fig:crc_tsne}
\end{figure}

\noindent CRC's effectiveness relies on our information-gap principle (F3): latent-space pruning is a better negative sampling strategy than pixel-level masking. We verify this using t-SNE visualizations \cite{maaten2008visualizing} of hidden states (Figure~\ref{fig:crc_tsne}). Across all four MLLMs, a clear pattern emerges: our pruned-token negative samples (Our\_Neg1/2/3) produce representations that cluster \textit{close} to the original image's representations (Ori\_IMG). This suggests our method is an \textit{in-distribution} probe. In contrast, masked image (Masked\_IMG) representations are often in completely different, distant clusters, suggesting a noisy, \textit{out-of-distribution} perturbation. This indicates that our CRC method uses a cleaner, more relevant signal to isolate bias, enabling a more precise calibration.

\textbf{4.4 Analysis over Hyperparameters}

\begin{figure}[t]
    \centering
    \includegraphics[width=1\linewidth]{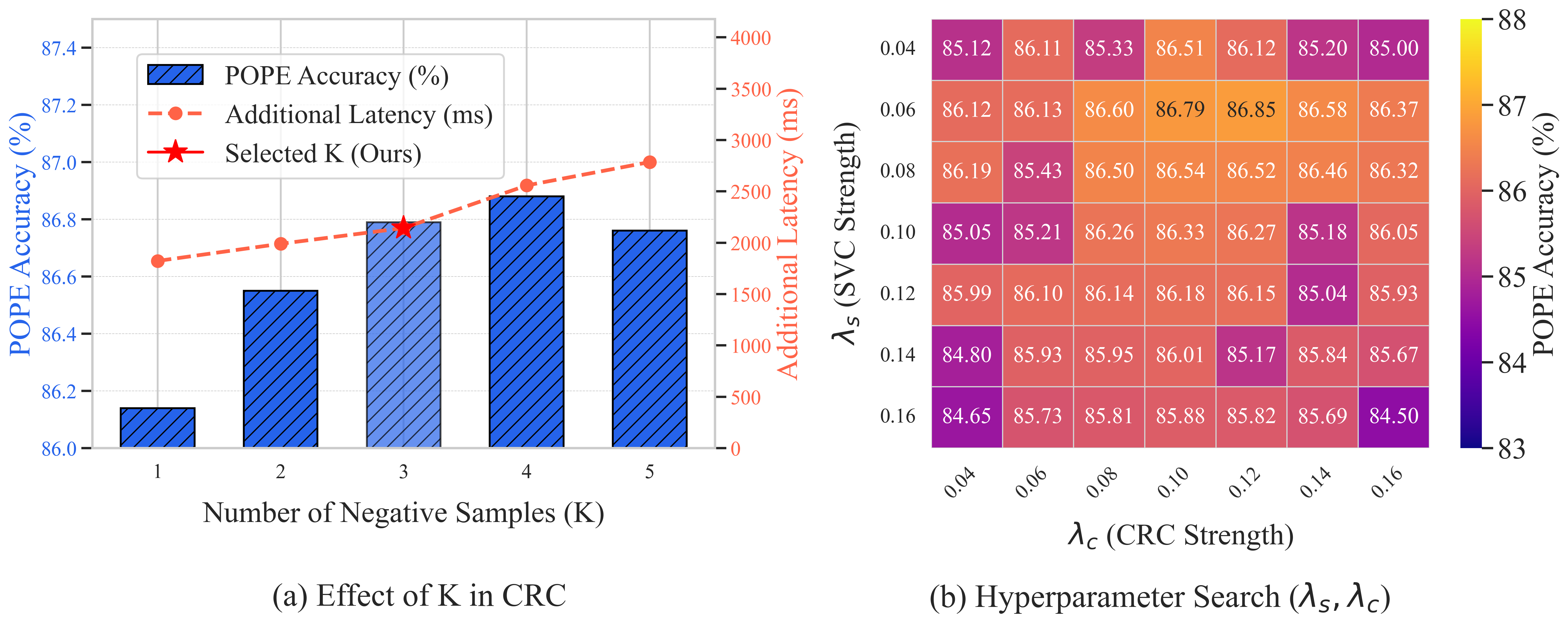} % Assumes file is named hyperparameter.png and in fig/ subdirectory
    \caption{
        \textbf{Hyperparameter Analysis.} \textbf{(a)} POPE Accuracy vs. Latency for different $K$ in CRC. We choose $K=3$ (red star) as the best balance of performance and speed. \textbf{(b)} Heatmap for SVC ($\lambda_s$) and CRC ($\lambda_c$) strength. Performance is robust, with an optimal region around $\lambda_s=0.06$ and $\lambda_c=0.10$.
    }
    \label{fig:hyperparameters}
\end{figure}

\noindent \textbf{Choice of K.} $K$ impacts both accuracy and speed. As shown in Figure~\ref{fig:hyperparameters}(a), performance peaks at $K=3$. While more samples add some stability, they also add extra latency. We thus select $K=3$ as the best trade-off.

\noindent \textbf{Calibration Strengths $\lambda_s$ and $\lambda_c$.} The heatmap in Figure~\ref{fig:hyperparameters}(b) shows our grid search results. Performance is stable across a wide range of values, confirming our method is not overly sensitive to these hyperparameters. We selected $\lambda_s = 0.06$ and $\lambda_c = 0.10$ as our default, as this region provides relatively strong results.

\begin{figure}[h]
    \centering
    \includegraphics[width=1\linewidth]{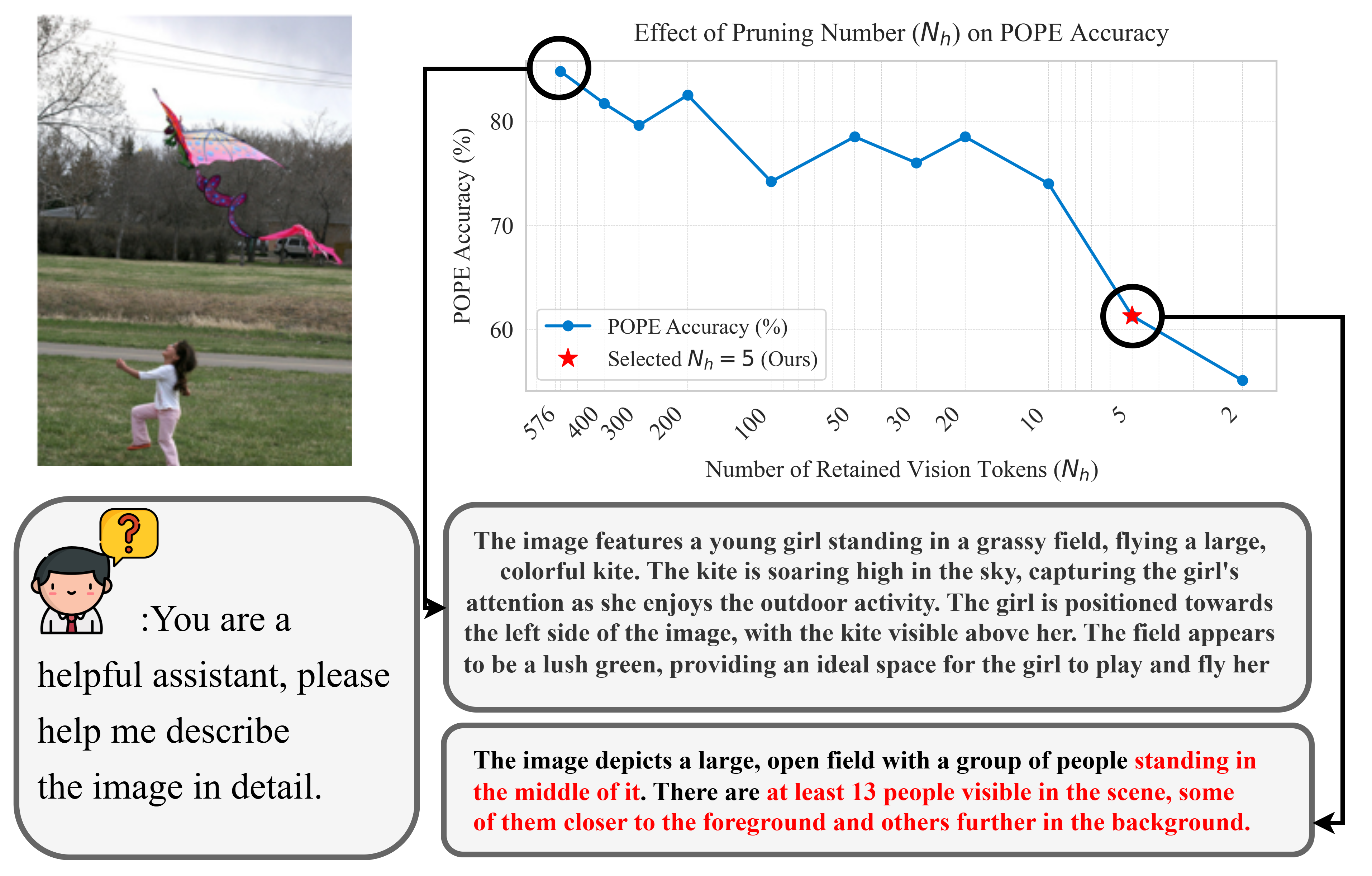} % Assumes this path is correct
    \caption{Effect of the number of retained vision tokens ($N_h$) on POPE Accuracy (LLaVA-1.5, COCO). The red star indicates our chosen value $N_h=5$. Accuracy drops sharply only when fewer than 20 tokens are retained.}
    \label{fig:ablation_pruning}
\end{figure}

\noindent \textbf{Choice of $N_h$.}
Unlike model pruning to accelerate decoding speed \cite{chen2024image,fan2025mathcal,wang2025sparsemm}, our goal is to \textit{reduce} visual grounding to induce a measurable hallucination signal. Our study (Figure~\ref{fig:ablation_pruning}) shows LLaVA-1.5's POPE accuracy is remarkably robust to random pruning, remaining stable even at $N_h=50$ (from 576). A noticeable decline begins only below $N_h=20$, with a sharp drop at $N_h=5$. This confirms $N_h \ge 20$ provides too much visual information, failing to trigger the bias we aim to probe. The sharp drop at $N_h=5$ indicates this is an effective operating point to force reliance on internal biases, so we select $N_h=5$ as our default.

%----------- FIGURE for Pruning Ratio -----------
%---------------------------------------------
\section{Related Work}
\label{sec:related}
Existing training-free methods for mitigating hallucination in MLLMs fall into two categories: \textit{vision enhancement} and \textit{text inertia suppression}. The former boosts visual signals during generation: e.g., \cite{liu2024paying} amplifies image token attention; \cite{huang2024opera} applies an over-trust penalty with retrospection allocation; \cite{tang2025seeing} uses an attention register to retain focus on relevant tokens; \cite{zoulook} re-injects visual tokens via FFN at the middle trigger layer. However, these attention-centric approaches often overlook the LLM’s strong \textit{text inertia}, leading to visual neglect despite enhanced attention. The latter category employs contrastive decoding to counter linguistic biases: \cite{leng2024mitigating} uses logits from a degraded image for calibration; \cite{wan2025only} amplifies key text based on a text-vision entropy ratio; \cite{zhangself} generates a synthetic image from an initial response for contrastive refinement. Yet these methods rely on negative samples from an image-level \textit{modality gap}, which, as shown in our analysis (Figure~\ref{fig:crc_tsne}, Table~\ref{tab:ablation_study}), introduces noise and yields unstable, out-of-distribution outputs.

In contrast, our framework observes this core challenge, unifies enhancement and suppression by treating vision tokens as the core cross-modal bridge. It jointly addresses vision-language imbalance through diverse visual cues (via SVC) and an information-gap–based negative sampling strategy (via CRC), grounded in the MLLM’s architectural principles and achieve superior performance.

\section{Conclusion}
\label{sec:conclusion}

In this work, we tackle MLLM hallucination by challenging the disjointed paradigm of training-free solutions, proposing a unified framework that repurposes vision tokens for dual complementary roles: enhancement and suppression. Leveraging systematic analyses of visual fading, semantic complementarity and latent negative sampling, our framework integrates two synergistic modules Synergistic Visual Calibration for robust visual grounding and Causal Representation Calibration for bias purification. Extensive experiments confirm our approach achieves SOTA MLLM hallucination mitigation with excellent inference efficiency.
% In this work, we addressed the critical issue of hallucination in MLLMs by challenging the prevailing disjointed paradigm of training-free solutions. We proposed a novel, unified framework built upon the principle of repurposing vision tokens for two complementary fates: enhancement and suppression. Grounded in systematic analyses of visual fading, semantic complementarity, and latent-level negative sampling, our framework introduces two synergistic modules: Synergistic Visual Calibration (SVC) for robust visual grounding and Causal Representation Calibration (CRC) for precise bias purification. Extensive experiments demonstrate that our unified approach achieves best MLLM hallucination mitigation, while maintaining excellent efficiency.
% \begin{align}
% H_t^{\prime(L_c)} &= (1 - \lambda_s) \cdot H_t^{(L_c)} + \lambda_s \cdot C_t \\
% h_{crc} &= h_{norm} + \lambda_c \cdot v_{nom}
% \end{align}

\section{Acknowledgement}
\label{sec:acknow}
This work was supported by National Natural Science Foundation of China Project (62536005, 62192783, 624B2063, 62506162) and Jiangsu Science and Technology Project (BF2025061, BG2024031, BK20251241).
{
    \small
    \bibliographystyle{ieeenat_fullname}
    \bibliography{main}
}


\end{document}